\documentclass[runningheads]{llncs}

\usepackage{accv}

\usepackage{accvabbrv}

\usepackage{graphicx}
\usepackage{booktabs}

\usepackage[accsupp]{axessibility}  
\usepackage{booktabs}
\usepackage{multirow}
\usepackage{tabularx}
\usepackage{colortbl}

\usepackage{hyperref}

\usepackage{orcidlink}

\begin{document}

\title{Self-Consistency Geometry Injection for Multi-view Consistent Image Generation} 

\titlerunning{Self-Consistency Geometry Injection}

\author{Yupeng Xiao\inst{1} \and
Yankai Wu\inst{1} \and
Jianjie Luo\inst{1}\thanks{Corresponding author: jianjieluo@gdut.edu.cn.} \and
Jiajie Lin\inst{2} \and \\
Yi Yu\inst{2} \and
Zhenguo Yang\inst{1}}

\authorrunning{Y.~Xiao et al.}

\institute{
Guangdong University of Technology, China
\and
Hiroshima University, Japan
}

\maketitle

\begin{abstract}
  Recent diffusion models have achieved impressive progress in single image to multi-view generation by synthesizing consistent novel views efficiently. Existing methods usually explicitly or implicitly couple geometry inference and texture synthesis, making generation vulnerable to inaccurate geometry prediction. Although several geometry-guided methods have explored decoupling, they usually treat predicted geometry as a fixed condition and do not explicitly model whether the geometry is reliable. We therefore propose Reliability-Guided Multi-View Generation (RGMV), which comprises a parameter-free Self-Consistency Geometry Gate (SCG) and a Reliability-aware Geometry Injection (RGI) module. Specifically, a geometry proposal generator produces multiple geometry predictions for the same reference image and target viewpoints under different stochastic sampling configurations. SCG measures their spatial agreement to derive a reliability map, which RGI uses to modulate the injection strength of geometry-conditioned residuals in the RGB synthesis model. Moreover, RGMV can adapt to different geometry proposal generators. As a result, RGMV preserves stronger geometric guidance in stable regions while attenuating unreliable geometry cues before they propagate into multi-view RGB generation. Extensive experiments show that reliability-aware geometry injection improves the geometric quality of multi-view generation. Code: \url{https://github.com/WISLab-GDUT/RGMV}.
\end{abstract}

\section{Introduction}

Multi-view Consistent Image Generation aims to synthesize visually coherent images of an object from novel viewpoints given only one reference image. Recent diffusion-based approaches have substantially improved the realism and efficiency of multi-view synthesis~\cite{xiao2025omnigen,shi2023mvdream,sauer2024fast,ho2020denoising,ramesh2022hierarchical,peebles2023scalable,ye2023ip,mou2024t2i,deng2023mv,gao2024cat3d} by leveraging large-scale image priors and joint view generation~\cite{wen2025ouroboros3d,qian2023magic123,luo2024unleashing,xiang2025structured,liu2024one,ye2023consistent}. Nevertheless, a single reference image provides only partial observations of the underlying object. The geometry of occluded and unseen regions therefore remains underconstrained, making it difficult to preserve plausible object structure and multi-view consistent local appearance.

\begin{figure}[t]
    \centering
    \includegraphics[width=\columnwidth]{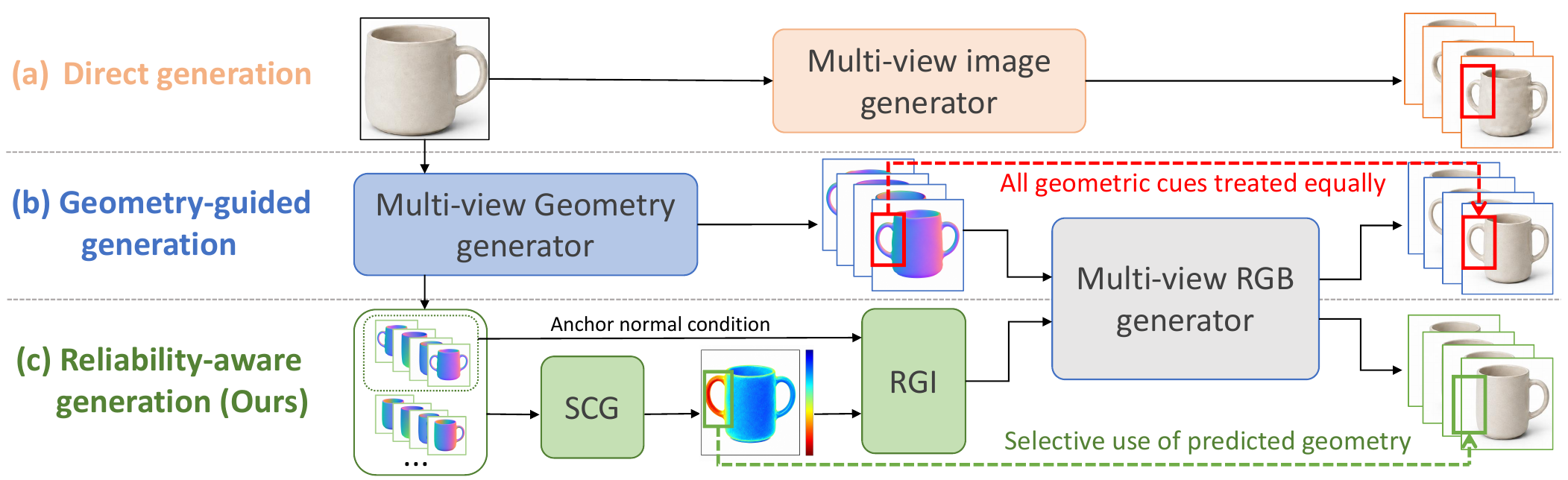}
    \caption{Comparison of three paradigms for multi-view image generation. 
(a) Direct generation entangles geometric reasoning and texture synthesis in a single generator, making geometric errors difficult to isolate. 
(b) Geometry-guided generation separates geometry prediction from RGB synthesis, but applies predicted geometry with spatially uniform influence. 
(c) RGMV conditions RGB synthesis on an anchor geometry prediction and uses per-view spatial reliability maps estimated by SCG to modulate geometry residuals through RGI.}
    \label{fig:motivation}
\end{figure}

To reduce this ambiguity, geometry-guided methods~\cite{long2024wonder3d,lu2024direct2,nguyen2026pointmap,liu2026calitex,fu2024geowizard,ke2024repurposing,ye2024stablenormal} introduce normal maps, depth maps, or other structural representations as explicit conditions for multi-view RGB synthesis. When the geometry condition is accurate, it provides view-aware structural cues that help constrain object shape and appearance in unseen regions. Figure~\ref{fig:motivation}(a,b) highlights two alternative formulations for multi-view generation. Direct methods do not expose geometry as an independently controllable signal, whereas geometry-guided pipelines expose it without accounting for its non-uniform reliability. However, geometry predicted from a single reference image is not uniformly reliable. Ambiguity in occluded regions, thin structures, and local boundaries can lead to spatially and view-wise varying errors in the predicted geometry. Existing geometry-conditioned pipelines~\cite{zhang2023adding,mou2024t2i,long2024wonder3d,lu2024direct2,nguyen2026pointmap} typically inject predicted geometry with a globally fixed strength, implicitly treating all geometric cues as equally trustworthy. This design creates an unavoidable trade-off that weak conditioning underuses reliable geometry, whereas strong conditioning propagates local geometric errors into RGB synthesis. Because the reliability of predicted geometry varies across spatial locations and target views, no single global injection scale can simultaneously preserve trustworthy structural cues and suppress unreliable ones. This limitation motivates a selective geometry-conditioning mechanism that preserves stable geometric cues while suppressing unreliable ones.

Therefore, we propose RGMV, a reliability-guided framework for geometry-guided multi-view generation. Figure~\ref{fig:motivation}(c) summarizes the resulting alternative. RGMV retains the decoupled pipeline but selectively controls the influence of predicted geometry rather than modifying the geometry itself. RGMV obtains multiple normal predictions for the same reference image and target poses from a geometry proposal generator under different stochastic seeds. SCG converts their agreement into spatial reliability maps, and RGI uses these maps to modulate the geometry residuals injected into the multi-view RGB synthesis model. This design preserves stronger geometry guidance in stable regions while limiting the influence of inconsistent predictions. Experiments show that SCG-guided RGI improves over spatially uniform injection with a frozen RGB synthesis model, and parameter-efficient adaptation further improves the full RGMV framework.

The main contributions of this work are as follows:
\begin{itemize}
\item We propose RGMV, which couples a parameter-free Self-Consistency Geometry Gate (SCG) for estimating spatial reliability from stochastic geometry agreement with a Reliability-aware Geometry Injection (RGI) module for modulating normal-conditioned residuals.
\item We introduce a parameter-efficient adaptation scheme for the RGI-equipped RGB synthesis model, which uses ground-truth normals during training and SCG-derived spatial variation only at inference, without dependence on a particular geometry proposal generator.
\item Extensive experiments on Objaverse and GSO demonstrate that RGMV consistently improves multiple geometry proposal generators in both multi-view RGB synthesis and downstream 3D reconstruction. Qualitative results show more stable structures on rendered objects and real captured images.
\end{itemize}

\section{Related Work}

\subsection{Multi-view Consistent Image Generation}

Diffusion-based models~\cite{liu2023zero,shi2023zero123++,liu2023syncdreamer,shi2023mvdream,voleti2024sv3d,NEURIPS2023_a0da690a} have become a major approach for synthesizing novel views from a single reference image. Zero-1-to-3~\cite{liu2023zero} introduces camera-conditioned diffusion, enabling novel-view generation under specified relative camera transformations. SyncDreamer~\cite{liu2023syncdreamer} further models the joint distribution of multiple target views and synchronizes their denoising process through cross-view feature interaction to improve multi-view consistency. SV3D~\cite{voleti2024sv3d} adapts latent video diffusion to orbital multi-view synthesis and incorporates explicit camera control for image-conditioned view generation. Era3D~\cite{li2024era3d} improves high-resolution multi-view generation through camera prediction and efficient row-wise attention that exploits epipolar correspondence across views. More recently, MV-Adapter~\cite{huang2025mv} introduces a lightweight adapter-based design that equips pretrained text-to-image diffusion models with multi-view generation capability while preserving the original backbone. These methods\cite{NEURIPS2023_a0da690a,tang2024mvdiffusion++,zheng2024free3d,gao2024cat3d,cao2025mvgenmaster} improve the realism, camera controllability, and cross-view coherence of image-based multi-view synthesis. Their primary focus is on constructing stronger multi-view generation priors or more effective cross-view interactions. In this work, we study how external geometry predicted from a single reference image should be used within an existing multi-view diffusion pipeline when the reliability varies across spatial regions and target views.

\subsection{Geometry-conditioned Diffusion Models}

Geometry conditions have been widely used to provide explicit spatial control for diffusion models\cite{zhang2023adding,mou2024t2i,ye2023ip,hu2024mvd,luo2023semantic,xu2025improving,luo2026keyid,cheng2025mvpaint}. ControlNet~\cite{zhang2023adding} introduces an additional trainable branch that injects spatial conditions, such as edges, depth maps, and normal maps, into a pretrained diffusion backbone. T2I-Adapter~\cite{mou2024t2i} instead learns lightweight adapters for external structural signals while keeping the original text-to-image model frozen. These methods establish effective mechanisms for incorporating geometry into image diffusion. Several works further incorporate geometry into multi-view or 3D generation\cite{long2024wonder3d,lu2024direct2,fu2024geowizard,ke2024repurposing,liu2026calitex}. Wonder3D~\cite{long2024wonder3d} jointly generates multi-view RGB images and normal maps through cross-domain diffusion and multi-view cross-domain attention. Direct2.5~\cite{lu2024direct2} models multi-view normal maps as a 2.5D representation and uses normal-conditioned multi-view image generation for appearance synthesis. PointmapDiff~\cite{nguyen2026pointmap} uses predicted pointmaps as geometric conditions for single-image novel-view synthesis through reference attention and ControlNet-based conditioning. In a related texture-generation setting, CaliTex~\cite{liu2026calitex} calibrates attention with known 3D structure to improve view-coherent texture synthesis. These methods demonstrate that explicit geometry can substantially improve view synthesis and appearance generation. However, they generally use the available geometric representation directly or condition generation with a globally fixed mechanism\cite{zhang2023adding,long2024wonder3d,lu2024direct2,nguyen2026pointmap}. Regarding this issue, RGMV uses multi-sample agreement to adapt the strength of geometry-conditioned residual injection at inference time.

\subsection{Consistency-based Reliability Estimation}

Multi-sample consistency has been explored as an inference-time signal for assessing the reliability of generative outputs\cite{luo2025exploring,jazbec2025generative,jang2026self}. Generative Uncertainty in Diffusion Models~\cite{jazbec2025generative} estimates the uncertainty of diffusion samples to identify potentially low-quality generations. Self-Refining Video Sampling~\cite{jang2026self} uses self-consistency to identify regions that benefit from selective refinement during video generation. These studies show that repeated stochastic generations can provide useful evidence about output stability without requiring an additional manually annotated reliability label. RGMV evaluates the stability of the geometry condition before it is injected into downstream multi-view RGB synthesis. Specifically, SCG measures agreement among normal predictions generated under different stochastic seeds and converts it into a spatial reliability map. This parameter-free signal directly modulates normal-conditioned residuals at inference time, allowing the multi-view diffusion model to retain stable geometric cues while attenuating geometry conditions that are locally inconsistent.

\begin{figure}[t]
    \centering
    \includegraphics[width=\columnwidth]{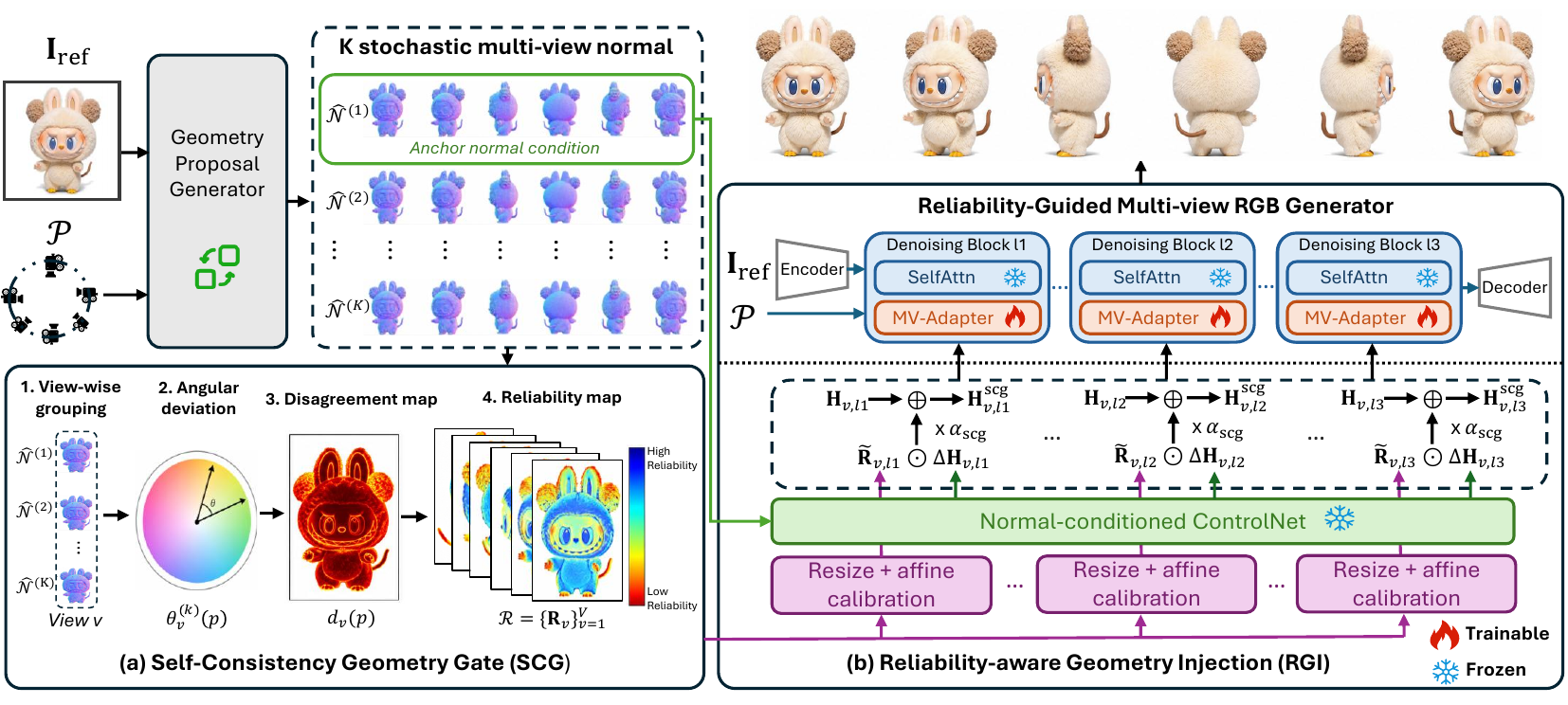}
    \caption{\textbf{Overview of RGMV.} Given a reference image and target camera poses, an off-the-shelf geometry proposal generator produces $K$ stochastic sets of multi-view normal predictions. The first prediction set serves as the anchor normal condition to produce multi-level geometry residuals. In parallel, Self-Consistency Geometry Gate (SCG) in (a) groups normal predictions by target view, measures their per-pixel angular disagreement, and converts it into per-view spatial reliability maps. Reliability-aware Geometry Injection (RGI) module in (b) resizes and affinely calibrates the corresponding reliability map at each denoising level $l$ to obtain $\widetilde{\mathbf R}_{v,l}$, which selectively modulates the geometry residual before it is injected into the multi-view diffusion backbone.}
    \label{fig:framework}
\end{figure}

\section{Method}
\label{sec:method}

\subsection{Overview}
\label{sec:overview}

Given a reference image $\mathbf{I}_{\mathrm{ref}}$ and a set of target camera poses $\mathcal{P}=\left\{\mathbf{P}_v\right\}_{v=1}^{V}$, our goal is to synthesize a set of multi-view RGB images $\hat{\mathcal{I}}=\left\{\hat{\mathbf{I}}_v\right\}_{v=1}^{V}$ that are consistent with the reference appearance and the target viewpoints. As illustrated in Fig.~\ref{fig:framework}, RGMV couples a geometry proposal generator with a multi-view RGB synthesis model equipped with the proposed SCG and RGI modules. In this work, we instantiate the geometry representation with multi-view normal maps.

The geometry proposal generator produces $K$ sets of normal predictions under different random seeds:
\begin{equation}
\hat{\mathcal{N}}^{(k)}
=
\left\{\hat{\mathbf{N}}_v^{(k)}\right\}_{v=1}^{V}
=
\mathcal{G}\left(
\mathbf{I}_{\mathrm{ref}},
\mathcal{P};
 s_k
\right),
\quad k=1,\ldots,K,
\label{eq:multi_geometry_prediction}
\end{equation}
where $\mathcal{G}$ denotes the geometry proposal generator and $s_k$ is the $k$-th random seed. We designate one prediction, $\hat{\mathcal{N}}^{(1)}$, as the anchor geometry condition for the RGB synthesis model, while the full prediction set is used to estimate its spatial reliability. Instead of treating the anchor normal maps as uniformly reliable, the proposed \emph{Self-Consistency Geometry Gate} (SCG) measures agreement among the $K$ predictions and derives a set of spatial reliability maps $\mathcal{R}=\left\{\mathbf{R}_v\right\}_{v=1}^{V}$.

The proposed \emph{Reliability-aware Geometry Injection} (RGI) module does not modify the normal maps themselves. Rather, it uses these reliability maps to control the strength with which normal-conditioned residuals are injected into the denoising network of the RGB synthesis model. The overall reliability-aware synthesis process can thus be written as:
\begin{equation}
\hat{\mathcal{I}}
=
\mathcal{T}\left(
\mathbf{I}_{\mathrm{ref}},
\mathcal{P},
\hat{\mathcal{N}}^{(1)},
\mathcal{R}
\right),
\label{eq:rgmv_overview}
\end{equation}
where $\mathcal{T}$ denotes the reliability-aware multi-view RGB synthesis process. Through RGI, RGMV preserves stronger geometry guidance in high-agreement regions and attenuates the corresponding geometry residuals in low-agreement regions. SCG is parameter-free and can be directly paired with a frozen RGB synthesis model through RGI, while the full RGMV further adapts the RGB synthesis model with ground-truth normal conditions through the same RGI pathway. The following sections describe the construction of the self-consistency reliability map and its use in reliability-aware geometry injection.

\subsection{Self-Consistency Geometry Gate}
\label{sec:scg}

Predicted normal maps can be reliable in some regions while remaining unstable in others. To estimate this variation without reliability supervision, SCG (Fig.~\ref{fig:framework}(a)) measures agreement among multiple normal predictions generated from the same reference image and target poses. At inference time, we sample the geometry proposal generator $K$ times with different random seeds.

For the $v$-th target view and pixel location $p$, let $\hat{\mathbf{n}}_v^{(k)}(p)\in\mathbb{R}^{3}$ denote the normal vector from the $k$-th prediction. We $\ell_2$-normalize each normal vector before aggregation and compute the mean normal direction as
\begin{equation}
\bar{\mathbf{n}}_v(p)
=
\frac{
\sum_{k=1}^{K}\hat{\mathbf{n}}_v^{(k)}(p)
}{
\left\lVert
\sum_{k=1}^{K}\hat{\mathbf{n}}_v^{(k)}(p)
\right\rVert_2+\varepsilon
},
\label{eq:mean_normal}
\end{equation}
where $\varepsilon$ is a small constant for numerical stability. We then measure disagreement by the mean angular deviation from the mean direction:
\begin{equation}
d_v(p)
=
\frac{1}{K}
\sum_{k=1}^{K}
\arccos\left(
\operatorname{clip}\left(
\left\langle
\hat{\mathbf{n}}_v^{(k)}(p),
\bar{\mathbf{n}}_v(p)
\right\rangle,
-1,
1
\right)
\right),
\label{eq:angular_deviation}
\end{equation}
where $\operatorname{clip}(\cdot,-1,1)$ restricts the inner product to the valid range of $\arccos(\cdot)$. We convert the disagreement map into a raw agreement-based reliability response:
\begin{equation}
\mathbf{R}_v(p)
=
\exp\left(-\gamma d_v(p)\right),
\label{eq:reliability_map}
\end{equation}
where $\gamma>0$ controls the sensitivity of the response to prediction disagreement.

The resulting reliability map is an agreement-based proxy for the stability of predicted geometry. It is not supervised by ground-truth normal errors and does not assume that prediction agreement exactly measures geometric accuracy. Instead, it provides a spatial signal for selectively controlling the influence of predicted geometry during RGB synthesis. SCG produces spatially varying reliability maps only at inference time from multiple predicted normal maps.

\subsection{Reliability-aware Geometry Injection}
\label{sec:injection}

The proposed RGI module (Fig.~\ref{fig:framework}(b)) injects geometry into the denoising network of the RGB synthesis model through residuals produced by a normal-conditioned ControlNet~\cite{zhang2023adding}. For notational clarity, let $\mathbf{H}_{v,l}$ denote the feature map for target view $v$ at the $l$-th denoising block, and let $\Delta\mathbf{H}_{v,l}$ denote the corresponding geometry residual. Conventional geometry-guided synthesis applies a spatially uniform injection strength:
\begin{equation}
\mathbf{H}^{\mathrm{uni}}_{v,l}
=
\mathbf{H}_{v,l}
+
\alpha_{\mathrm{uni}}
\Delta\mathbf{H}_{v,l},
\label{eq:uniform_injection}
\end{equation}
where $\alpha_{\mathrm{uni}}$ is a fixed geometry injection scale. This formulation treats all spatial locations of the predicted geometry as equally reliable.

In contrast, RGI modulates geometry residuals using the reliability map derived by SCG in Sec.~\ref{sec:scg}. For each denoising block, we resize the reliability map to match the spatial resolution of $\mathbf{H}_{v,l}$ and apply an affine calibration:
\begin{equation}
\widetilde{\mathbf{R}}_{v,l}
=
r_{\min}
+
\left(r_{\max}-r_{\min}\right)
\mathcal{U}_{l}\left(\mathbf{R}_v\right),
\label{eq:reliability_calibration}
\end{equation}
where $\mathcal{U}_{l}(\cdot)$ denotes spatial resizing to the resolution of $\mathbf{H}_{v,l}$. The parameters $r_{\min}$ and $r_{\max}$ specify an affine calibration of the raw reliability response. They are not clipping bounds, and Eq.~\ref{eq:reliability_calibration} does not normalize the gate to a prescribed value range. The single-channel map $\widetilde{\mathbf{R}}_{v,l}$ is broadcast along the feature-channel dimension before multiplication. The reliability-aware feature update is
\begin{equation}
\mathbf{H}^{\mathrm{scg}}_{v,l}
=
\mathbf{H}_{v,l}
+
\alpha_{\mathrm{scg}}
\widetilde{\mathbf{R}}_{v,l}
\odot
\Delta\mathbf{H}_{v,l},
\label{eq:scg_injection}
\end{equation}
where $\odot$ denotes element-wise multiplication and $\alpha_{\mathrm{scg}}$ is the base injection scale. Thus, RGI uses the SCG-derived reliability map to preserve stronger geometry residuals in high-agreement regions and attenuate residuals in regions with low geometric agreement. Uniform geometry injection in Eq.~\ref{eq:uniform_injection} is a special case of Eq.~\ref{eq:scg_injection} with $\widetilde{\mathbf{R}}_{v,l}=\mathbf{1}$.

We consider uniform and RGMV adaptation settings. Both use ground-truth normal maps: the former applies the constant scale $\alpha_{\mathrm{uni}}$, whereas the latter follows the RGI update in Eq.~\ref{eq:scg_injection} with base scale $\alpha_{\mathrm{scg}}$. For RGMV, $K$ identical normal copies yield a raw reliability map of one and thus a constant calibrated gate of $r_{\max}$. Training therefore retains the RGI pathway without spatial variation and remains independent of the geometry proposal generator. At inference, ground-truth normals are replaced by the anchor predicted normals, and SCG derives spatially varying reliability maps from the $K$ predicted sets.

During RGB synthesis model adaptation, we keep the normal-conditioned ControlNet~\cite{zhang2023adding} frozen and optimize the MV-Adapter parameters with the standard diffusion denoising objective:
\begin{equation}
\mathcal{L}_{\mathrm{denoise}}
=
\mathbb{E}_{z_0,t,\boldsymbol{\epsilon}}
\left[
\left\lVert
\boldsymbol{\epsilon}
-
\boldsymbol{\epsilon}_{\theta}
\left(
 z_t,
 t,
 \mathbf{I}_{\mathrm{ref}},
 \mathcal{P};
 \mathcal{N}_{\mathrm{cond}},
 \mathcal{R}_{\mathrm{cond}}
\right)
\right\rVert_2^2
\right],
\label{eq:denoising_loss}
\end{equation}
where $z_t$ is the noisy latent of the clean target latent $z_0$ at timestep $t$, $\boldsymbol{\epsilon}$ is the sampled Gaussian noise, and $\boldsymbol{\epsilon}_{\theta}$ denotes the denoising network of the RGB synthesis model. During RGMV fine-tuning, $\mathcal{N}_{\mathrm{cond}}$ is the ground-truth normal condition and $\mathcal{R}_{\mathrm{cond}}$ is the spatially constant reliability condition induced by its identical copies. During inference, they are replaced by $\hat{\mathcal{N}}^{(1)}$ and the SCG reliability maps computed from multiple geometry predictions, respectively, which are supplied to RGI for residual modulation. The uniform baseline follows the same objective but applies the update in Eq.~\ref{eq:uniform_injection} without a reliability condition.

\section{Experiments}
\subsection{Dataset and Metric}
We train RGMV on a curated Objaverse~\cite{deitke2023objaverse} subset of approximately $100$K objects, with more than $60$K for training and the remainder for validation. Ablations use a fixed $100$-object validation subset, and  the Google Scanned Objects (GSO)~\cite{downs2022google} provides a fixed $100$-object test set. Both of them are selected with seed $42$. For 2D synthesis, we report PSNR, SSIM, and LPIPS~\cite{zhang2018unreasonable}, which measure pixel fidelity, structural similarity, and perceptual similarity respectively. For 3D reconstruction, lower Chamfer Distance (CD) and higher Volume IoU indicate better geometric quality.
\subsection{Implementation Details}
\label{sec:implementation_details}
Our multi-view RGB synthesis model is built on Stable Diffusion v2.1-base~\cite{rombach2022high}. We use MV-Adapter~\cite{huang2025mv} as the multi-view consistency module and a normal-conditioned ControlNet~\cite{zhang2023adding} for geometry residual injection. We use $K=3$, $\gamma=0.20$, $\alpha_{\mathrm{scg}}=0.08$, and $[r_{\min},r_{\max}]=[0,1]$, selected in Sec.~\ref{sec:ablation}. During training, we optimize only the parameters of MV-Adapter for $120{,}000$ steps with batch size $3$ and learning rate $1\times10^{-5}$, while freezing the diffusion backbone and ControlNet, and the Classifier-Free Guidance (CFG) scale is $3.0$. For each reference image, the proposal generator produces $K$ stochastic sets of six-view normal maps, and RGMV synthesizes the corresponding RGB views at $0^\circ$ (front), $45^\circ$ (front-right), $90^\circ$ (right), $180^\circ$ (back), $270^\circ$ (left), and $315^\circ$ (front-left). SCG assumes no specific camera angles, but arbitrary or continuous pose evaluation is beyond the scope of this work. Peak GPU memory is approximately 45\,GB for training. Unless specified otherwise, the default geometry proposal generator of RGMV is a separately trained MV-Adapter for multi-view normal generation, trained with the same optimization settings and parameter-efficient adaptation strategy as the RGB synthesis model.

\subsection{Performance Comparison}
\label{sec:performance_comparison}

\noindent\textbf{Compared Methods.}
We compare RGMV with two representative multi-view image generation baselines, Zero-1-to-3~\cite{liu2023zero} and SyncDreamer~\cite{liu2023syncdreamer}, as well as three recent systems with distinct multi-view generation designs, including Wonder3D++~\cite{yang2025wonder3d++}, MV-Adapter~\cite{huang2025mv}, and Era3D~\cite{li2024era3d}. Zero-1-to-3 performs camera-conditioned novel-view synthesis from a single reference image, whereas SyncDreamer~\cite{liu2023syncdreamer} explicitly models the joint distribution of multiple target views. Wonder3D++~\cite{yang2025wonder3d++}, MV-Adapter, and Era3D~\cite{li2024era3d} represent complementary approaches based on joint RGB--geometry generation, lightweight multi-view adaptation, and epipolar-aware multi-view generation, respectively. The latter three methods can additionally provide multi-view geometry proposals for RGMV. We therefore evaluate each of them both as a standalone multi-view generation method and as a geometry proposal generator integrated with RGMV.

\begin{table}[t]
    \centering
    \caption{Comparison of RGB synthesis and 3D reconstruction on GSO. All RGMV variants share the same adapted RGB synthesis model; parentheses indicate the geometry proposal generator. CD denotes Chamfer Distance, and IoU denotes Volume IoU.}
    \label{tab:gso_quantitative}

    \small
    \setlength{\tabcolsep}{3pt}
    \renewcommand{\arraystretch}{1.08}

    \begin{tabularx}{\linewidth}{
        @{}l
        *{5}{>{\centering\arraybackslash}X}
        @{}
    }
        \toprule
        \multicolumn{1}{c}{\multirow{2}{*}{Method}}
        & \multicolumn{3}{c}{Multi-view image generation}
        & \multicolumn{2}{c}{3D reconstruction} \\
        \cmidrule(lr){2-4}
        \cmidrule(lr){5-6}
        & PSNR$\uparrow$
        & SSIM$\uparrow$
        & LPIPS$\downarrow$
        & CD$\downarrow$
        & IoU$\uparrow$ \\
        \midrule

        Zero-1-to-3
        & 17.38 & 0.8479 & 0.1383 & 0.071 & 0.33 \\
        
        SyncDreamer
        & 18.83 & 0.8212 & 0.2146 & 0.066 & 0.39 \\

        Wonder3D++
        & 20.05 & 0.8575 & 0.1463 & 0.062 & 0.44 \\

        MV-Adapter
        & 20.13 & 0.8664 & 0.1268 & 0.036 & 0.54 \\
        
        Era3D
        & 20.77 & 0.8634 & 0.1253 & 0.038 & 0.53 \\

        \midrule

        RGMV (Wonder3D++)
        & 21.21 & 0.8736 & 0.1195 & 0.037 & 0.54 \\

        RGMV (MV-Adapter)
        & 21.38
        & 0.8729
        & 0.1157
        & \textbf{0.026}
        & \textbf{0.62} \\
        
        RGMV (Era3D)
        & \textbf{21.94}
        & \textbf{0.8737}
        & \textbf{0.1139}
        & 0.028
        & 0.60 \\

        \bottomrule
    \end{tabularx}
\end{table}

\noindent\textbf{Quantitative Comparison.}
Table~\ref{tab:gso_quantitative} compares multi-view image generation and downstream 3D reconstruction on GSO~\cite{downs2022google}. Among the standalone methods, Era3D achieves the highest PSNR and lowest LPIPS, whereas MV-Adapter provides the highest SSIM. RGMV improves PSNR by $1.16$, $1.25$, and $1.17$~dB when using geometry proposals from Wonder3D++~\cite{yang2025wonder3d++}, MV-Adapter~\cite{huang2025mv}, and Era3D~\cite{li2024era3d}, respectively, with corresponding improvements in SSIM and LPIPS. All RGMV variants share the same adapted RGB synthesis model, supporting compatibility with different geometry sources. Downstream reconstruction using Trellis~\cite{xiang2025structured} yields lower Chamfer Distance and higher Volume IoU than the corresponding standalone method for all three geometry sources. RGMV using Era3D proposals achieves the best multi-view image generation results, while RGMV using MV-Adapter proposals achieves the lowest Chamfer Distance of $0.026$ and the highest Volume IoU of $0.62$. The specific contribution of SCG-guided RGI is examined through controlled ablations in Sec.~\ref{sec:ablation}.
\begin{figure}[t]
    \centering
    \includegraphics[width=\columnwidth]{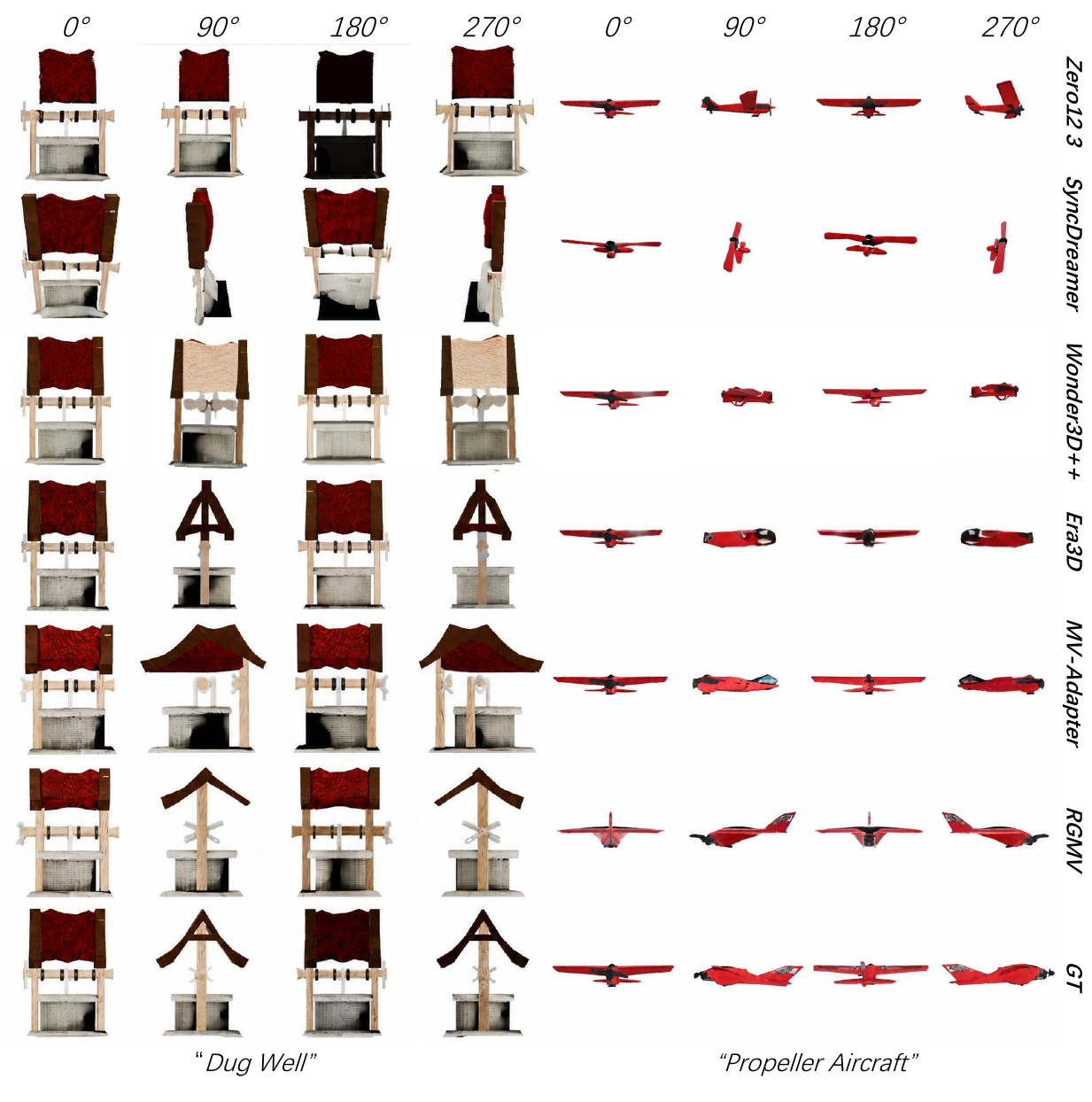}
    \caption{Qualitative comparison on rendered objects with ground-truth target views. Each row shows the multi-view results of one method under the same target viewpoints. }
    \label{fig:rendered}
\end{figure}
\begin{figure}[t]
    \centering
    \includegraphics[width=\columnwidth]{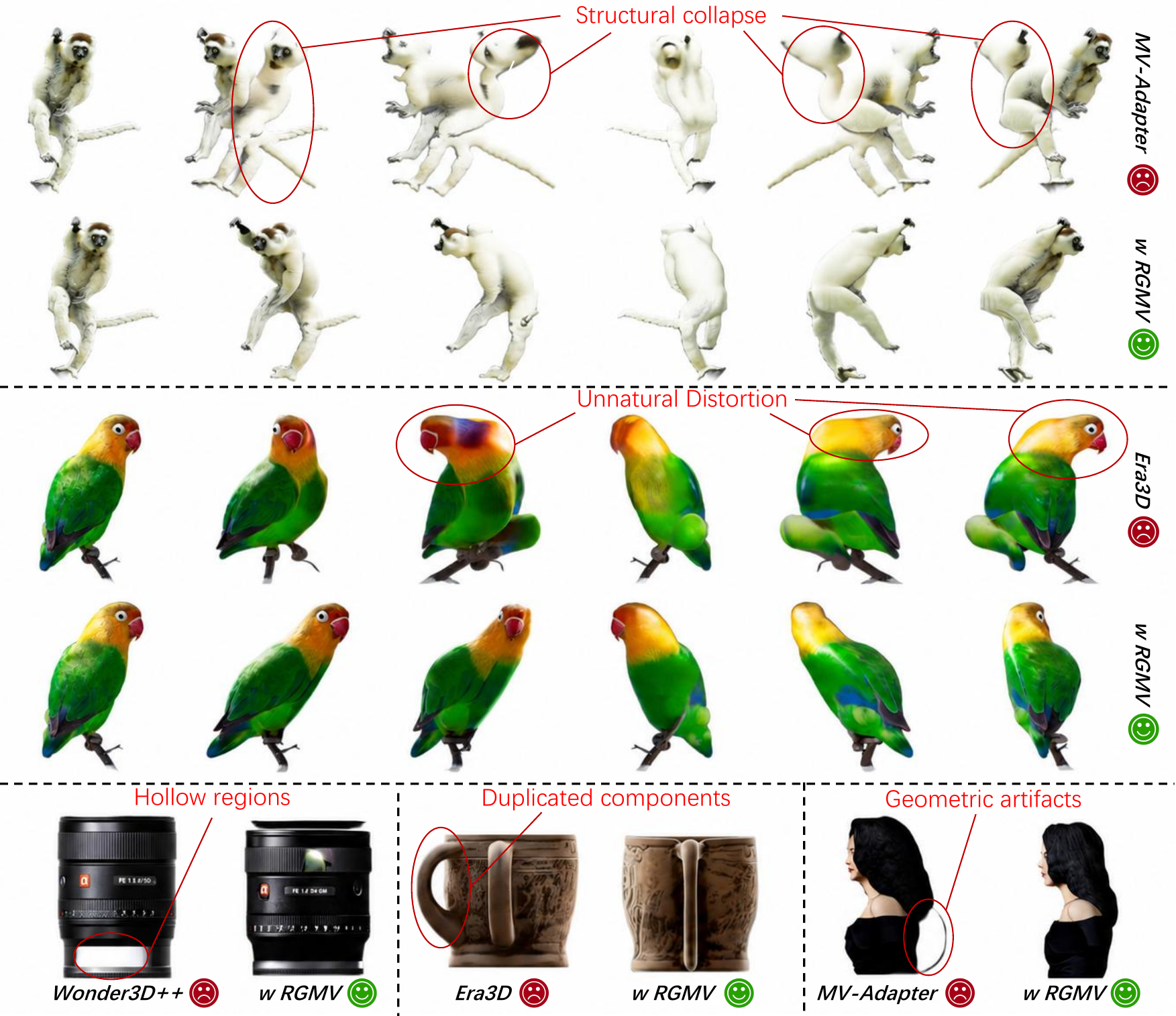}
    \caption{Qualitative comparison on real captured images without ground-truth target views. The examples are organized by representative failure modes commonly observed in multi-view generation. Each pair compares a baseline result with its corresponding RGMV result under the same input and target viewpoints.}
    \label{fig:real}
\end{figure}

\noindent\textbf{Qualitative Comparison.}
Figure~\ref{fig:rendered} compares generated views of held-out rendered objects with ground-truth targets. For ``Dug Well'' and ``Propeller Aircraft'', RGMV produces more faithful object structures and local details across viewpoints, bringing the generated views closer to ground truth. Figure~\ref{fig:real} compares results on real captured images without ground-truth target views. RGMV alleviates structural collapse, distortions, hollow regions, duplicated components, and other geometry artifacts. The red circles mark the affected baseline regions. In an audit of a fixed random sample of $50$ real images, the baseline (MV-Adapter~\cite{huang2025mv}) exhibits geometry-related failures in $18/50$ cases ($36.0\%$). RGMV fixes or alleviates $14$ of these $18$ failures ($77.8\%$), while $4$ remain unresolved. It also introduces a new geometry artifact in $1/50$ cases ($2.0\%$). Together, the rendered and real-image examples indicate that RGMV improves the use of predicted geometry in both controlled evaluation settings and unconstrained real-world inputs.
\subsection{Ablation Study}
\label{sec:ablation}

We conduct controlled ablations on the fixed $100$-object Objaverse~\cite{deitke2023objaverse} validation subset to isolate the effects of predicted-normal conditioning, SCG-guided reliability-aware geometry injection, and RGB synthesis model adaptation. All geometry-conditioned variants use multi-view normal proposals from our separately trained MV-Adapter. The ``Pure i2mv'' row in Table~\ref{tab:ablation_main} corresponds to the image-only MV-Adapter RGB synthesis model and is evaluated without normal-conditioned residual injection. All geometry-conditioned variants retain the same MV-Adapter RGB backbone and are evaluated under the same reference images, target viewpoints, and generation protocol. Because RGI operationalizes the spatial reliability maps estimated by SCG, we evaluate their joint contribution by replacing the SCG-derived spatial gate with uniform injection rather than treating RGI as a separately removable component. The adapted variants follow the same training configuration described in Sec.~\ref{sec:implementation_details}.

\noindent\textbf{Validity of the Reliability Signal.}
To assess whether the SCG reliability map is a useful proxy for geometry reliability, we examine the association between angular disagreement $d$ across predicted normals and anchor-to-GT angular error $e$. With $K=3$, we evaluate over 30 million pixel locations using Spearman's rank correlation coefficient $\rho$ for correlation and AUROC for high-error discrimination. Table~\ref{tab:reliability_validation} shows that $d$ correlates positively with $e$, and SCG helps distinguish high-error pixels from lower-error ones.

\begin{table}[t]
    \centering
    \caption{Association between SCG reliability and ground-truth normal error on Objaverse with $K=3$.}
    \label{tab:reliability_validation}
    \small
    \setlength{\tabcolsep}{3pt}
    \renewcommand{\arraystretch}{1.08}

    \begin{tabularx}{\linewidth}{
        @{}
        >{\hsize=1.00\hsize\raggedright\arraybackslash}X
        >{\hsize=1.00\hsize\centering\arraybackslash}X
        >{\hsize=1.00\hsize\centering\arraybackslash}X
        @{}
    }
        \toprule

        \multicolumn{1}{c}{Statistic}
        & \multicolumn{1}{c}{Result}
        & \multicolumn{1}{c}{Range} \\

        \midrule

        Pooled Spearman $\rho$
        & 0.299
        & $[-1,1]$ \\

        Per-sample $\rho$ ($\mu/\mathrm{med}/\sigma$)
        & 0.266 / 0.289 / 0.166
        & $[-1,1]$ \\

        \arrayrulecolor[gray]{0.65}
        \midrule
        \arrayrulecolor{black}

        AUROC ($e>20^\circ/30^\circ$)
        & 0.729 / 0.710
        & $[0,1]$ \\

        \bottomrule
    \end{tabularx}
\end{table}

\begin{figure}[t]
    \centering
    \includegraphics[width=\columnwidth]{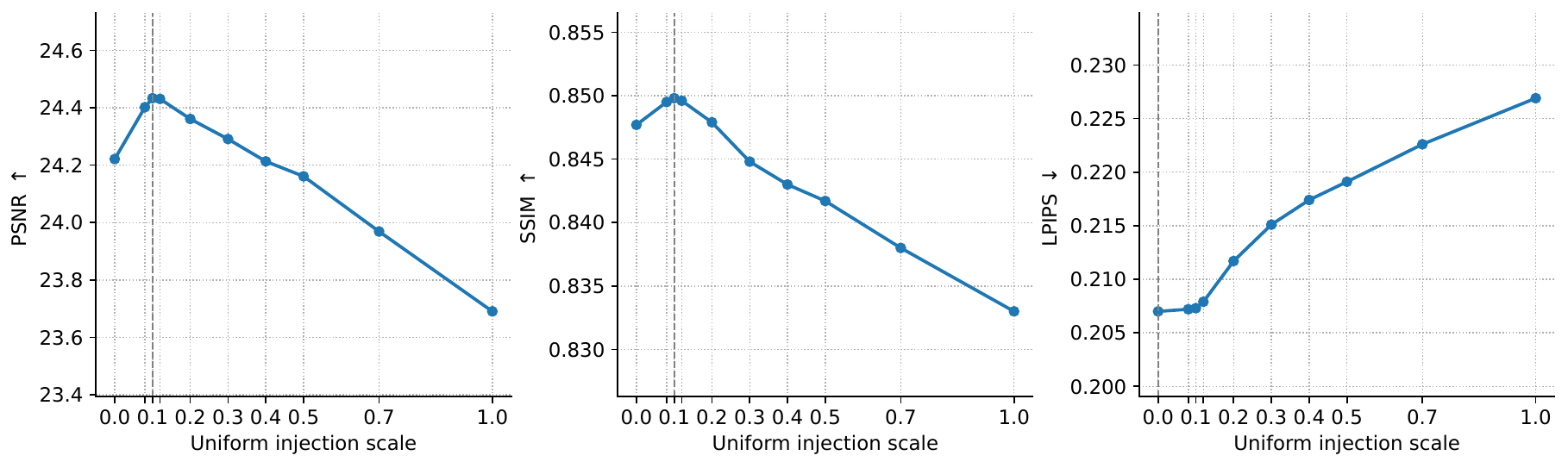}
    \caption{Sensitivity of spatially uniform geometry injection on Objaverse with a frozen RGB synthesis model. The dashed line marks $\alpha_{\mathrm{uni}}=0.1$, which is selected for the uniform-injection baseline. Larger global scales progressively degrade performance.}
    \label{fig:uniform_injection_scale}
\end{figure}

\noindent\textbf{Effect of SCG-guided RGI.}
With the RGB synthesis model frozen, adding spatially uniform normal-conditioned residuals to the image-only MV-Adapter baseline improves PSNR and SSIM, confirming that predicted geometry provides useful structural cues for multi-view generation. However, its LPIPS score does not improve, indicating that a fixed global injection strength cannot consistently translate predicted geometry into better perceptual synthesis. Replacing the uniform scalar with the SCG-derived reliability map in RGI improves all three metrics, yielding a further $0.20$~dB gain in PSNR, a $0.0047$ improvement in SSIM, and a $0.0138$ reduction in LPIPS. Since these variants share the same predicted normal maps, frozen MV-Adapter RGB synthesis model, and normal-conditioned residual pathway, the gain isolates the benefit of SCG-guided RGI over spatially uniform geometry injection. Figure~\ref{fig:uniform_injection_scale} shows that $\alpha_{\mathrm{uni}}=0.1$ achieves the highest PSNR and SSIM among the tested uniform scales and is therefore used in Table~\ref{tab:ablation_main}. At a matched mean gate of $0.9138$, SCG outperforms the constant and best Sobel-edge controls on all three metrics, with PSNR of $24.63$ versus $24.52$ and $24.54$, respectively. Uniform injection remains a global compromise across regions with different geometry reliability, whereas SCG-guided RGI performs spatially varying modulation at the pixel level. 

\begin{table*}[t]
    \centering
    \caption{Controlled ablation of predicted geometry conditioning, reliability-aware injection, and multi-view RGB synthesis model adaptation. ``Uniform'' denotes spatially constant geometry residual injection with $\alpha_{\mathrm{uni}}=0.1$, selected by the sweep in Fig.~\ref{fig:uniform_injection_scale}. }
    \label{tab:ablation_main}
    \small
    \setlength{\tabcolsep}{4.5pt}
    \renewcommand{\arraystretch}{1.12}

    \begin{tabular*}{\linewidth}{@{\extracolsep{\fill}}lccrrr@{}}
        \toprule
        \multicolumn{1}{c}{Setting}
        & Injection
        & RGB Model
        & PSNR$\uparrow$
        & SSIM$\uparrow$
        & LPIPS$\downarrow$ \\
        \midrule

        Pure i2mv & -- & Frozen & 24.22 & 0.8477 & 0.2069 \\
        \addlinespace[1pt]

        Uniform Injection & Uniform & Frozen & 24.43 & 0.8498 & 0.2073 \\
        SCG Injection & SCG & Frozen & 24.63 & 0.8545 & 0.1935 \\
        \addlinespace[1pt]

        Uniform Injection & Uniform & Adapted & 25.14 & 0.8569 & 0.1948 \\
        \textbf{RGMV} & SCG & Adapted & \textbf{25.31} & \textbf{0.8621} & \textbf{0.1823} \\

        \bottomrule
    \end{tabular*}
\end{table*}

\begin{table*}[t]
    \centering
    \caption{Hyperparameter ablation on Objaverse. Each panel varies one factor while keeping the others at the defaults in Sec.~\ref{sec:implementation_details}.}
    \label{tab:scg_hyperparams}
    \small
    \renewcommand{\arraystretch}{1.08}
    \newcommand{\paneltitlegap}{1.5pt}
    \begin{minipage}[t]{0.48\textwidth}
        \centering
        \textbf{(a) Reliability coefficient $\gamma$}\\[\paneltitlegap]
        \setlength{\tabcolsep}{5.5pt}
        \begin{tabular}{crrr}
            \toprule
            Setting & PSNR$\uparrow$ & SSIM$\uparrow$ & LPIPS$\downarrow$ \\
            \midrule
            0.02 & 24.61 & 0.8538 & 0.1950 \\
            0.05 & 24.61 & 0.8539 & 0.1948 \\
            0.10 & 24.62 & 0.8541 & 0.1942 \\
            \textbf{0.20} & \textbf{24.63} & \textbf{0.8545} & 0.1935 \\
            0.30 & 24.61 & 0.8543 & \textbf{0.1934} \\
            0.50 & 24.61 & 0.8541 & 0.1939 \\
            0.70 & 24.52 & 0.8535 & 0.1957 \\
            1.00 & 24.36 & 0.8495 & 0.2012 \\
            \bottomrule
        \end{tabular}

        \medskip
        \textbf{(c) Number of geometry predictions $K$}\\[\paneltitlegap]
        \setlength{\tabcolsep}{5.5pt}
        \begin{tabular}{crrr}
            \toprule
            Setting & PSNR$\uparrow$ & SSIM$\uparrow$ & LPIPS$\downarrow$ \\
            \midrule
            2 & 24.58 & 0.8540 & \textbf{0.1923} \\
            \textbf{3} & \textbf{24.63} & \textbf{0.8545} & 0.1935 \\
            4 & \textbf{24.63} & \textbf{0.8545} & 0.1935 \\
            5 & \textbf{24.63} & \textbf{0.8545} & 0.1938 \\
            \bottomrule
        \end{tabular}
    \end{minipage}
    \hfill
    \begin{minipage}[t]{0.48\textwidth}
        \centering
        \textbf{(b) Base scale $\alpha_{\mathrm{scg}}$}\\[\paneltitlegap]
        \setlength{\tabcolsep}{5.5pt}
        \begin{tabular}{crrr}
            \toprule
            Setting & PSNR$\uparrow$ & SSIM$\uparrow$ & LPIPS$\downarrow$ \\
            \midrule
            0.02 & 24.39 & 0.8490 & 0.1999 \\
            0.04 & 24.53 & 0.8519 & 0.1971 \\
            0.06 & 24.57 & 0.8529 & 0.1956 \\
            \textbf{0.08} & \textbf{24.63} & \textbf{0.8545} & \textbf{0.1935} \\
            0.10 & 24.61 & 0.8542 & 0.1946 \\
            0.15 & 24.46 & 0.8509 & 0.2027 \\
            0.20 & 24.40 & 0.8486 & 0.2092 \\
            0.30 & 24.31 & 0.8465 & 0.2158 \\
            \bottomrule
        \end{tabular}

        \medskip
        \textbf{(d) Affine calibration interval $[r_{\min},r_{\max}]$}\\[\paneltitlegap]
        \setlength{\tabcolsep}{5.0pt}
        \begin{tabular}{lrrr}
            \toprule
            Setting & PSNR$\uparrow$ & SSIM$\uparrow$ & LPIPS$\downarrow$ \\
            \midrule
            $[0.0,0.5]$ & 24.53 & 0.8529 & 0.1963 \\
            \textbf{$[0.0,1.0]$} & \textbf{24.63} & \textbf{0.8545} & \textbf{0.1935} \\
            $[0.2,0.8]$ & 24.60 & 0.8541 & 0.1941 \\
            $[0.0,2.0]$ & 24.41 & 0.8494 & 0.2080 \\
            \bottomrule
        \end{tabular}
    \end{minipage}
\end{table*}

\noindent\textbf{Effect of RGB Synthesis Model Adaptation.}
Table~\ref{tab:ablation_main} further evaluates the contribution of adapting the multi-view RGB synthesis model. Compared with the frozen uniform-injection variant, model adaptation improves PSNR from $24.43$ to $25.14$, SSIM from $0.8498$ to $0.8569$, and LPIPS from $0.2073$ to $0.1948$. This result shows that adaptation improves the compatibility between the RGB synthesis model and normal-conditioned residuals. More importantly, RGMV remains superior under the same adapted synthesis model. Replacing uniform injection with SCG-guided RGI improves PSNR by $0.17$~dB, SSIM by $0.0052$, and LPIPS by $0.0125$. Therefore, the gain of RGMV cannot be attributed only to model adaptation. Adaptation improves the model's ability to exploit geometry conditions, whereas SCG supplies spatial reliability maps and RGI uses them to determine how strongly the predicted geometry should influence generation at each spatial location. Their combination yields the best result.

\noindent\textbf{Sensitivity to SCG and RGI Hyperparameters.}
Table~\ref{tab:scg_hyperparams} evaluates the sensitivity to $\gamma$, $\alpha_{\mathrm{scg}}$, $K$, and $[r_{\min},r_{\max}]$. The default setting, $\gamma=0.20$, $\alpha_{\mathrm{scg}}=0.08$, $K=3$, and $[r_{\min},r_{\max}]=[0,1]$, provides the strongest overall trade-off across PSNR, SSIM, and LPIPS. The selected configuration avoids both overly aggressive suppression of geometry cues and insufficient discrimination between stable and unstable prediction regions. The results remain stable over a moderate range of $\gamma$, with $\gamma=0.20$ achieving the best structural metrics and nearby values producing comparable perceptual quality. This indicates that SCG does not require an excessively sharp reliability response. Large $\gamma$ instead suppresses useful geometry cues in moderately consistent regions. The base injection scale exhibits a clearer optimum at $\alpha_{\mathrm{scg}}=0.08$. Smaller values underuse reliable normal cues, whereas larger values amplify geometry-conditioned residuals beyond what reliability modulation can compensate for. Using $K=3$ geometry predictions improves PSNR and SSIM over $K=2$, although $K=2$ achieves slightly lower LPIPS. Increasing to $K=4$ yields only negligible quality changes, which are obscured by rounding in Table~\ref{tab:scg_hyperparams}, while increasing total runtime by approximately $23\%$ (Table~\ref{tab:runtime_by_k}). We therefore select $K=3$ as a trade-off between synthesis quality and inference cost. Finally, the affine calibration interval $[0,1]$ performs best overall, since a lower upper bound weakens reliable geometry guidance across high-agreement regions during RGB synthesis, whereas a larger one over-amplifies residual injection.

\noindent\textbf{Inference Cost.}
Table~\ref{tab:runtime_by_k} reports the runtimes of geometry proposal generation, SCG computation, and RGB synthesis on an RTX 4090 GPU. At $K=3$, runtime is $16.65$ seconds, approximately $1.88\times$ that at $K=1$. SCG accounts for approximately $0.7\%$ of runtime. Repeated geometry proposals dominate the overhead. Increasing $K$ from $3$ to $4$ raises runtime by approximately $23\%$.

\begin{table}[t]
    \centering
    \caption{Inference time on an RTX 4090 GPU for different geometry proposal sets $K$.}
    \label{tab:runtime_by_k}
    \small
    \begin{tabular*}{\linewidth}{@{\extracolsep{\fill}}ccccc@{}}
        \toprule
        $K$ & proposal generation (s) & SCG computation (s) & RGB synthesis (s) & Total (s) \\
        \midrule
        1 & $3.84 \pm 0.02$ & $0.053 \pm 0.001$ & $4.95 \pm 0.01$ & $8.85 \pm 0.02$ \\
        2 & $7.71 \pm 0.02$ & $0.089 \pm 0.002$ & $4.96 \pm 0.01$ & $12.76 \pm 0.02$ \\
        3 & $11.56 \pm 0.02$ & $0.124 \pm 0.003$ & $4.97 \pm 0.01$ & $16.65 \pm 0.02$ \\
        4 & $15.41 \pm 0.02$ & $0.163 \pm 0.002$ & $4.96 \pm 0.01$ & $20.54 \pm 0.03$ \\
        \bottomrule
    \end{tabular*}
\end{table}

\section{Conclusion}
In this paper, we presented RGMV, a reliability-guided framework for multi-view image generation. RGMV considers geometry predicted from a single reference image useful but not uniformly reliable across spatial regions and target views. The proposed SCG is parameter-free and operates at inference time to construct a reliability map from the agreement among multiple stochastic geometry predictions, while RGI uses this map to selectively modulate geometry residuals during RGB synthesis. Experiments demonstrate that RGMV improves multiple geometry proposal generators on both multi-view image synthesis and downstream 3D reconstruction. Qualitative results further show that reliability-guided conditioning reduces geometry-related artifacts and produces more stable object structures across viewpoints.

%
%
\bibliographystyle{splncs04}
\bibliography{main}
\end{document}